\documentclass[10pt,twocolumn,letterpaper]{article}

\usepackage{booktabs, tabularx, wrapfig, adjustbox, multirow} %
\usepackage{adjustbox} %
\usepackage{graphicx} %
\usepackage{microtype}

\usepackage{algorithm}
\usepackage{algpseudocode}
\usepackage{wrapfig}
\usepackage{graphicx}

\newif\ifarxiv

\arxivtrue      %

\ifarxiv
  \usepackage{animate} %
\fi

\newcommand{\teaservid}{%
  \ifarxiv
    \animategraphics[width=\linewidth,autoplay,loop,keepaspectratio]{12}%
      {images/teaser_vid_vertical/frame_0}{000}{225}%
  \else
    \includegraphics[width=\linewidth,keepaspectratio]{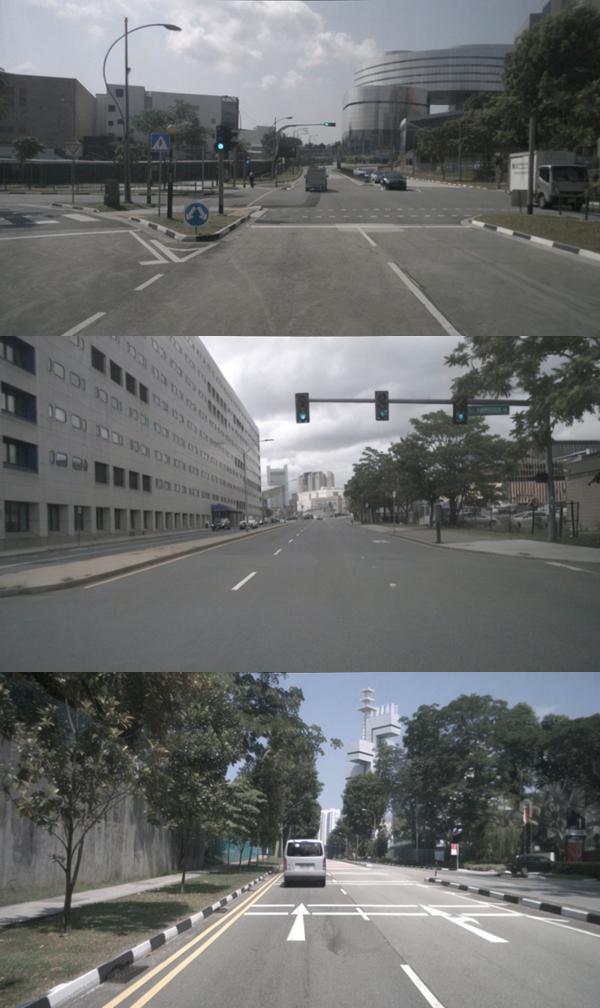}%
  \fi
}

\newcolumntype{L}[1]{>{\raggedright\arraybackslash}p{#1}}
\newcolumntype{C}[1]{>{\centering\arraybackslash}p{#1}}
\newcolumntype{R}[1]{>{\raggedleft\arraybackslash}p{#1}}

\newcommand\mypara[1]{\noindent\textbf{#1.}}

\newcommand{\ourmethod}{AutoScape\xspace}
\newcommand{\ourtitle}{AutoScape: Geometry-Consistent Long-Horizon Scene Generation}

\newcommand{\BB}[1]{}

\ifarxiv
  \usepackage[pagenumbers]{iccv} %
\else
  \usepackage{iccv} %
\fi

\definecolor{iccvblue}{rgb}{0.21,0.49,0.74}
\usepackage[pagebackref,breaklinks,colorlinks,allcolors=iccvblue]{hyperref}

\def\paperID{} %
\def\confName{ICCV}
\def\confYear{2025}

\title{\ourtitle}

\author{%
Jiacheng Chen$^{*\,2}$ \quad 
Ziyu Jiang$^{*\,\dagger\,1}$ \quad 
Mingfu Liang$^{3}$ \quad 
Bingbing Zhuang$^{1}$ \quad \\
Jong-Chyi Su$^{1}$ \quad 
Sparsh Garg$^{1}$ \quad 
Ying Wu$^{3}$ \quad 
Manmohan Chandraker$^{1,4}$ \\ \\
$^{1}$NEC Labs America \quad
$^{2}$Simon Fraser University \quad
$^{3}$Northwestern University \quad 
$^{4}$UC San Diego }

\begin{document}

\makeatletter
\let\@oldmaketitle\@maketitle
\renewcommand{\@maketitle}{\@oldmaketitle
\vspace*{-1.8em}
\centerline{%
\begin{minipage}{0.8\textwidth}
  \centering
  \includegraphics[width=\textwidth]{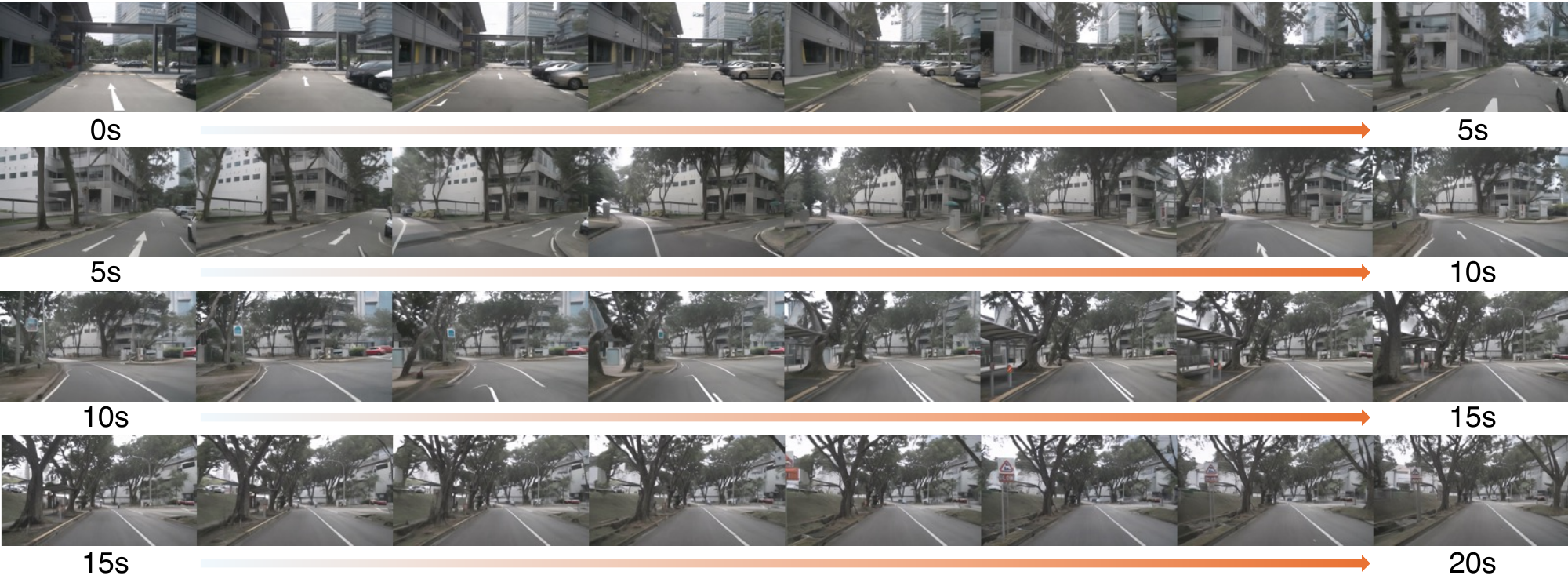}
\end{minipage}
\begin{minipage}{0.17\textwidth}
  \centering
  \begin{tabular}{c}
    \teaservid\\
  \end{tabular}
\end{minipage}%
}
\vspace*{-0.8em}
\captionof{figure}{\ourmethod generates long-horizon and 3D-consistent driving scenes from a single input image, producing high-quality videos over a temporal span of 20 seconds. \textcolor{magenta}{Click} the images in the right column within Adobe Reader to play three example videos.}
\label{fig:teaser}
\vspace*{1em}
}
\makeatother
\maketitle

\begingroup
\renewcommand\thefootnote{}\footnotetext{\textsuperscript{*}Equal Contribution. $\dagger$ Project Lead.}
\addtocounter{footnote}{-1}
\endgroup

\begin{abstract}
This paper proposes \ourmethod, a long-horizon driving scene generation framework. At its core is a novel RGB-D diffusion model that iteratively generates sparse, geometrically consistent keyframes, serving as reliable anchors for the scene's appearance and geometry. To maintain long-range geometric consistency, the model 1) jointly handles image and depth in a shared latent space, 2) explicitly conditions on the existing scene geometry (\ie, rendered point clouds) from previously generated keyframes, and 3) steers the sampling process with a warp-consistent guidance. 
Given high-quality RGB-D keyframes, a video diffusion model then interpolates between them to produce dense and coherent video frames.   
\ourmethod generates realistic and geometrically consistent driving videos of over 20 seconds, improving the long-horizon FID and FVD scores over the prior state-of-the-art by 48.6\% and 43.0\%, respectively. Project page: \url{https://auto-scape.github.io}.
\end{abstract}

\section{Introduction}
\label{sec:intro}

Recent advances in high-quality video generation are rapidly transforming various applications, ranging from robotics to mixed reality, where the synthesis of realistic visual data is crucial. A particular example is autonomous driving, where the photorealistic generation of driving videos plays a crucial role in simulation and verification. But despite promising prospects, generating 3D driving scenes that remain coherent and consistent over long horizons remains a fundamental challenge. Current generative methods, while achieving impressive photorealism \cite{fridman2023scenescape,hollein2023text2room,gao2024magicdrive3d,zhao2024drivedreamer4d,yu2024wonderjourney,yu2024wonderworld}, often struggle with maintaining physical realism \cite{gao2024magicdrive3d} and suffer from quality degradation during auto-regressive generation \cite{fridman2023scenescape,deng2024streetscapes}, making spatiotemporal coherence over long horizons a critical unsolved challenge.

We introduce \textbf{\ourmethod}, a novel framework that addresses the challenges of long-horizon 3D-consistent driving scene generation by leveraging explicit geometry awareness. Given the observation that degrading geometric consistency is the key bottleneck of long-horizon scene or video generation, we decompose the problem hierarchically into \textit{sparse RGB-D keyframe generation} and \textit{dense video interpolation}. The core idea is to train a powerful RGB-D diffusion model to generate highly consistent keyframes, which serve as reliable anchors for the scene's global appearance and geometry, robustly handling a large span. Given the reliable anchors, a video diffusion model then interpolates between keyframes, refining rendered point clouds into a coherent video.

Our RGB-D diffusion model generates high-quality keyframes with three key designs:
{\em 1) joint RGB-D modeling}, which operates on the joint distribution of color and depth for more coherent appearance and geometry, and conducts pre-training on large-scale paired data from diverse sources beyond driving to obtain general RGB-D priors;
{\em 2) explicit geometry conditioning}, where each generated RGB-D keyframe is directly conditioned on the existing scene's appearance and geometry, in the format of rendered point clouds;
{\em 3) warp consistent guidance}, a classifier guidance style approach that steers the diffusion model's sampling process toward better geometric alignment with previous keyframes, mitigating the accumulation of errors throughout long-term generation.

Compared to those methods that handle spatial and temporal consistency using only the temporal modules of a video diffusion model~\cite{gao2024magicdrive3d,gao2024vista}, our hierarchical approach offers greater robustness in terms of long-horizon generation, since the RGB-D diffusion model first produces sparse yet highly consistent keyframes as global anchors rather than directly generating dense frames.
Compared to existing works that also employ explicit 3D modeling and produce keyframes~\cite{yu2024wonderjourney}, our method demonstrates superior keyframe quality by the joint RGB-D modeling, geometry conditioning, and warp-consistent guidance.

As shown in \autoref{fig:teaser}, \ourmethod generates long-horizon, 3D-consistent, and high-quality scenes with a video duration of 20 seconds containing 250 frames. Quantitatively, it achieves significant improvements over the previous state-of-the-art method, with reductions of 48.6\% and 43.0\% in FID and FVD scores, respectively, in terms of long-horizon video generation. To summarize, our contributions are threefold:

\begin{itemize}
\item \textbf{\ourmethod}, a novel framework that jointly generates the appearance and geometry of long-range driving scenes using a hierarchical approach of keyframe generation and interpolation.

\item A new RGB-D diffusion model featuring geometry-aware conditioning and guidance to enforce long-range 3D consistency, ensuring both geometric stability and high-fidelity visual quality.

\item State-of-the-art quantitative and qualitative results in long-horizon driving scene generation as demonstrated by comprehensive experiments.
\end{itemize}

\section{Related Works}
\label{sec:related}

\mypara{Diffusion Models}
Diffusion-based generative models~\cite{ho2020ddpm,song2020diffusion_sde,sohl2015dm_early} have fueled a surge in generative AI. While the theoretical advancements keep improving the mathematical formulation, sampling speed, and generation quality~\cite{song2020ddim,karras2022EDM,lipman2022flow-matching,liu2022rectified-flow,song2023consistency-models}, Variants of diffusion models have extended the early success in image generation~\cite{rombach2022ldm,ramesh2022dalle2,dhariwal2021dmbeatsgan} to a broad spectrum, including video~\cite{blattmann2023svd,blattmann2023alignlatents_video,wu2023tune-a-video}, audio~\cite{kong2020diffwave,liu2023audioldm}, 3D~\cite{poole2022dreamfusion,tang2024mvdiffusion++,luo2021diffusion_pc,cheng2023sdfusion,Zeng2022LIONLP,lin2024drive,kong2024eschernet,li20244k4dgen}, motion synthesis~\cite{zhang2024motiondiffuse,jiang2023motiondiffuser}, visual editing~\cite{brooks2023instructpix2pix,wei2024omniedit,wu2024neural,michel2023object3dit,chen2025blenderfusion}, and more.
We employ diffusion models for RGB-D keyframe generation and interpolation toward long-horizon driving scene generation.

\mypara{3D Scene Generation}
Diffusion models have been widely applied in scene generation. One prominent framework is iterative inpainting, where scenes are progressively expanded from an initial image, like SceneScape~\cite{fridman2023scenescape}, Text2Room~\cite{hollein2023text2room}, and WonderJourney~\cite{yu2024wonderjourney}. WonderWorld~\cite{yu2024wonderworld} advances this paradigm by enhancing 3D consistency and supporting interactive user control. Another widely adopted paradigm involves generating 360-degree panoramas that can be converted into 3D models~\cite{Tang2023MVDiffusion,dreamscene360,fang2023ctrl-room,schult2024controlroom3d}. Additional methods focus on producing high-level scene layouts~\cite{xu2023diffscene,shabani2023housediffusion,lu2024scenecontrol} or generating LiDAR point clouds to represent 3D scene structures~\cite{ran2024lidar-diffusion,zhang2023copilot4d}. In autonomous driving, methods like MagicDrive3D~\cite{gao2024magicdrive3d} and DriveDreamer4D~\cite{zhao2024drivedreamer4d} generate driving videos using diffusion models, which are subsequently transformed into 3D scenes for efficient simulation. However, the temporal span of video diffusion models constrains these approaches in terms of the scene scale.

\mypara{Street View Generation} 
Although numerous studies investigated street‑view generation with reconstruction systems~\cite{yang2023unisim, sun2024lidarf,yan2024street, tonderski2024neurad,yang2023emernerf, chen2024omnire}, the recent advances in diffusion models have made generative simulation popular. Diffusion‑based approaches have been applied to sensor simulation or data augmentation, leveraging layout conditioning such as HD maps or object bounding boxes to produce realistic urban scenes~\cite{swerdlow2024BEVGen, liu2024mvpbev, yang2023bevcontrol, wang2023drivedreamer, gao2023magicdrive, wen2024panacea, hu2023gaia, russell2025gaia2}. MagicDrive~\cite{gao2023magicdrive} and MVPbev~\cite{liu2024mvpbev} employ cross‑view attention mechanisms to synthesize multi‑camera images. StreetScapes~\cite{deng2024streetscapes} proposes an autoregressive video diffusion model to generate long‑range street‑view videos, conditioned on 2.5D maps. Vista~\cite{gao2024vista} constructs a driving model capable of producing extended, high‑fidelity driving videos with action controls. DriveArena~\cite{yang2024drivearena} integrates components from previous methods to establish a generative closed‑loop simulator. Our work focuses on scene and video generation over substantially longer temporal horizons.

\mypara{Concurrent Works} 
Several concurrent efforts explore long-horizon street-view synthesis~\cite{gao2024magicdrivev2,lu2024infinicube}. InfiniCube~\cite{lu2024infinicube} constructs an explicit sparse-voxel 3D world to guide a video diffusion model, generating unbounded driving scenes. MagicDrive-V2~\cite{gao2024magicdrivev2} introduces an efficient video-diffusion architecture that scales to longer sequences. In contrast, our approach develops a novel RGB-D diffusion model that iteratively produces sparse, geometry-consistent keyframes, which in turn facilitate long-horizon video generation.

\begin{figure*}[!th]
  \centering
\includegraphics[width=\textwidth,
]{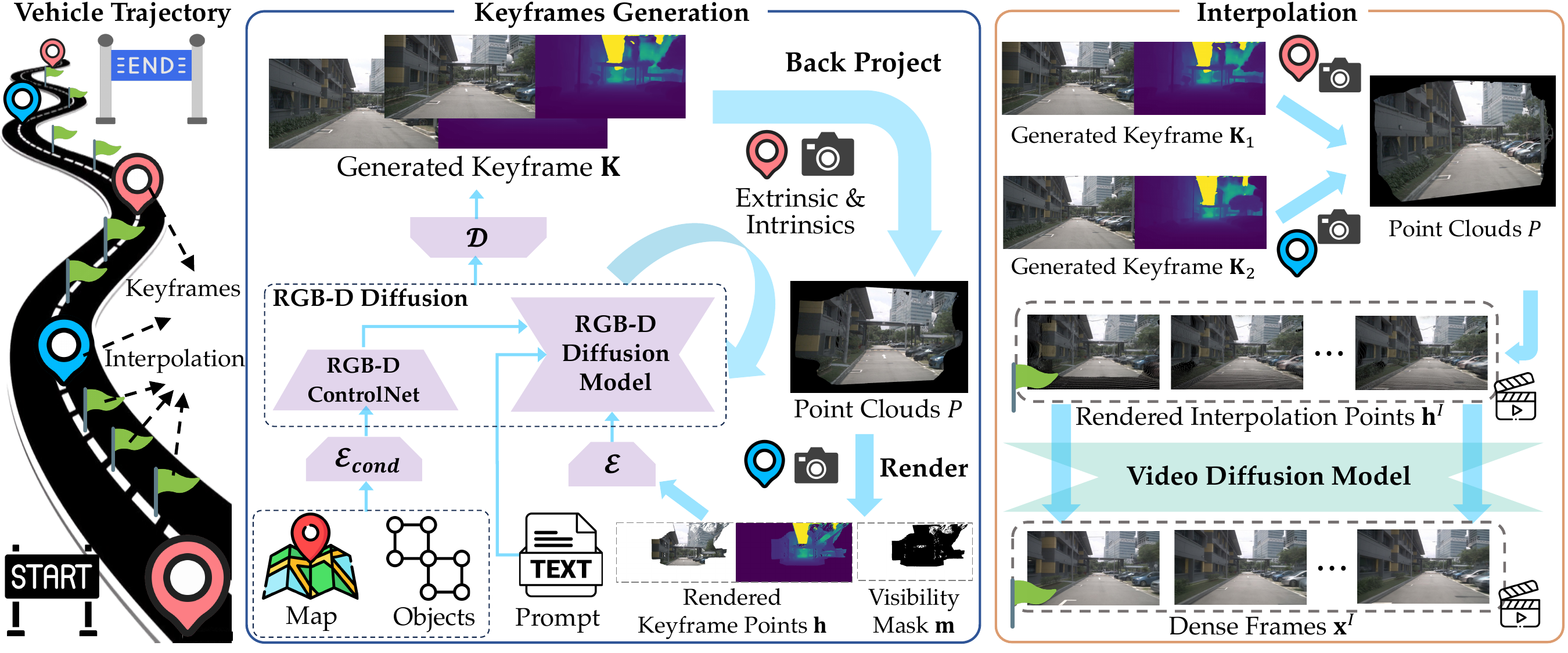}
\vspace{-1.5em}
\caption{Pipeline of the \ourmethod. The vehicle trajectory defines the location of keyframes and interpolation frames, spanning a long-horizon 3D space. The \textit{Keyframes Generation} stage iteratively generates geometrically consistent keyframes with an RGB-D diffusion model as global scene anchors.
The \textit{Interpolation} stage then produces dense frames with a video diffusion model. The keyframe viewpoints are indicated by \raisebox{-0.3mm}{\includegraphics[width=0.015\textwidth]{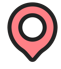}} and the interpolation viewpoints are marked by \raisebox{-0.3mm}{\includegraphics[width=0.015\textwidth]{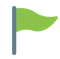}}, best viewed in color.
}
\label{fig:pipeline}
\end{figure*}

\section{Preliminary}
\label{sec:prelim}

Diffusion models lay the foundation of our scene generation framework. Diffusion-based generative models~\cite{sohl2015dm_early,ho2020ddpm,song2020diffusion_sde} have recently emerged as a dominant family of generative models, capable of capturing complex data distributions through iterative denoising processes. The core mechanism involves a pre-defined forward diffusion process \( q(\mathbf{x}_t \mid \mathbf{x}_{t-1}) \) that incrementally adds Gaussian noise to the data over \( T \) timesteps, transforming an original data sample \( \mathbf{x}_0 \) into a noisy \(\mathbf{x}_T\), defined by:
\begin{equation}
    q(\mathbf{x}_t \mid \mathbf{x}_{t-1}) = \mathcal{N}\left(\mathbf{x}_t; \sqrt{1 - \beta_t}\, \mathbf{x}_{t-1}, \beta_t \mathbf{I}\right)
\end{equation}
where \( \beta_t \) denotes the variance schedule controlling the noise level at each timestep, and \( \mathbf{I} \) is the identity matrix. The \emph{reverse process} aims to recover the original data by learning a parameterized denoising model \( p_\theta(\mathbf{x}_{t-1} \mid \mathbf{x}_t) \):
\begin{equation}
p_\theta(\mathbf{x}_{t-1} \mid \mathbf{x}_t) = \mathcal{N}\left(\mathbf{x}_{t-1}; \boldsymbol{\mu}_\theta(\mathbf{x}_t, t), \boldsymbol{\Sigma}_\theta(\mathbf{x}_t, t)\right)
\end{equation}
$\boldsymbol{\mu}_\theta $ and $ \boldsymbol{\Sigma}_\theta $ representing the mean and covariance functions modeled by neural networks with parameters $ \theta $. By iteratively applying this process starting from a Gaussian noise $\mathbf{x}_T$, the model generates new data samples that resemble the training data distribution.

Latent Diffusion Models (LDMs)~\cite{rombach2022ldm} operate in the compressed latent space of a pre-trained autoencoder rather than the raw high-dimensional data space, which enhances computational efficiency without compromising generative performance. Our method is based on the LDM formulation.

\section{\ourmethod}
\label{sec:method}

\subsection{Framework Overview}
\label{sec:overall_framework}

This paper focuses on generating long-horizon, high-quality, and 3D-consistent driving scenes. Although recent advancements in general video generation techniques~\cite{wu2023tune-a-video, blattmann2023svd} have made promising progress in producing driving videos~\cite{gao2023magicdrive, gao2024vista}, ensuring 3D consistency across hundreds of frames (\eg, over 20 seconds) remains hard. Long-range temporal and geometric consistency are the central challenges.

\ourmethod is a two-stage scene generation framework aiming for robust long-term coherency and stability (\autoref{fig:pipeline}). In the \textit{keyframe generation} stage, a RGB-D diffusion model jointly generates keyframes and the corresponding point clouds to anchor the scene’s global appearance and geometry. In the \textit{interpolation} stage, dense frames are first rendered from the consecutive RGB-D keyframes and then refined into coherent images using a video diffusion model. The explicit geometry modeling makes the first stage produce consistent yet sparse keyframes, which then serve as reliable conditions for the interpolation stage.

\mypara{Keyframe Generation Process} The keyframe generation process, illustrated in \autoref{fig:pipeline} (left), generates keyframes iteratively along specified sparse viewpoints. Each iteration comprises three steps: back-projection, rendering, and diffusion model generation. In the first iteration, we begin with a real input image or a generated RGB-D image, back-projecting the image into 3D space as point clouds $\mathcal{P}$ with camera parameters. 
The back projection is defined as:
\begin{equation}
    \mathcal{P} = \mathbf{B}(\mathcal{X}_{\text{rgb}}, \mathcal{X}_{\text{depth}}, \mathcal{C}),
\end{equation}
where
\(\mathcal{P} = \{ \mathbf{P}_i \}_{i=1}^N\) denotes the set of 3D point clouds obtained from the existing $N$ keyframes.
\(\mathcal{X}_{\text{rgb}} = \{ \mathbf{x}_{\text{rgb}, i} \}_{i=1}^N\) and \(\mathcal{X}_{\text{depth}} = \{ \mathbf{x}_{\text{depth}, i} \}_{i=1}^N\) represent the collections of RGB images and depth images of the keyframes, respectively.
\(\mathcal{C} = \{ c_i \}_{i=1}^N\) is the set of camera parameters (including intrinsics and extrinsics) corresponding to each keyframe.
\(\mathbf{B}(\cdot)\) is the back-projection function that reconstructs the 3D point clouds \(\mathbf{P}_i\) from the RGB and depth images using the camera parameters \(c_i\) for each keyframe \(i\).
Subsequently, the rendered keyframe points are produced by projecting the point clouds onto the image plane of the next keyframe:
\begin{equation}
    \mathbf{h}, \mathbf{m} = \mathbf{R}(\mathcal{P}, c),
    \label{equ:render-1}
\end{equation}
where \(\mathbf{h}\) is the rendered keyframe points, essentially a coarse image with noise and holes.
\(\mathbf{m}\) is the corresponding visibility mask indicating the presence of projected points.
\(\mathbf{R}(\cdot)\) is the rendering function that projects the 3D point clouds onto the image plane defined by the target camera parameters \(c\).

The rendered keyframe points \(\mathbf{h}\) and the visibility mask serve as the conditioning input for the RGB-D diffusion model, along with the map, object boxes, and prompt conditions. The generated RGB-D keyframe $\mathbf{K}$ would then contribute to the next iteration by adding its back-projection into the existing point clouds. The auto-regressive process runs in reverse along the trajectory, starting from the end and iteratively moving to the next nearest keyframe viewpoint until it reaches the start. \S~\ref{sec:rgbd_diff} presents the details of this RGB-D diffusion model.

Recent works, such as WonderJourney~\cite{yu2024wonderjourney,yu2024wonderworld}, have also explored the use of keyframes. However, these methods primarily rely on pretrained image inpainting models for keyframe generation. We propose a novel conditional RGB-D diffusion model, which offers key advantages in terms of \textbf{generalizability} and \textbf{geometry awareness}. Specifically, we introduce a two-stage training framework that enables the diffusion model to be pre-trained on large-scale RGB-D data (millions of images), thereby improving its \textbf{generalizability}. Moreover, previous methods predict depth solely from the generated RGB frame, which restricts the ability of different frames to access details from the previously generated geometry, often resulting in inconsistent depth maps. To overcome this limitation, we make the diffusion model conditioned on both appearance and geometry, thus generating more coherent keyframes. This design significantly enhances the model's \textbf{geometry awareness} and ensures better consistency with existing scene geometry.

\mypara{Warp Consistent Guidance} While the explicit geometry conditioning improves the long-term consistency, we still observe that the generated content of the RGB-D diffusion model occasionally misaligns with the rendered keyframe points from previous keyframes. To 
further improve the consistency at test time, we propose 
a warp consistent guidance mechanism, steering the sampling process toward better 3D alignment during inference. More details are in \S~\ref{sec:warpConsistent}

\mypara{Interpolation} 
The interpolation process generates dense frames by interpolating between two consecutive keyframes. The interpolation process is conditioned on the rendered 3D point clouds derived from the keyframes and utilizes an off-the-shelf video diffusion model from ViewCrafter~\cite{yu2024viewcrafter}. Existing approaches~\cite{yu2024wonderworld,yu2024wonderjourney} rely on pre-trained image generation models to interpolate sparse keyframes, neglecting the global consistency of the interpolated frames. 
Given high-quality keyframes generation, a point-cloud-conditioned video diffusion model can produce highly coherent interpolation along the rendering trajectory.

In the rest of the section, \S~\ref{sec:rgbd_diff}, \S~\ref{sec:warpConsistent}, and \S~\ref{sec:videodiff} elaborate on the RGB-D diffusion model, the warp consistent guidance mechanism, and the interpolation video diffusion models.

\subsection{RGB-D Diffusion for Keyframe Generation}
\label{sec:rgbd_diff}

The core of \ourmethod is a diffusion model that auto-regressively generates RGB-D keyframes. It incorporates robust geometric priors by explicitly modeling the depth and training on a large dataset with curated depth information. The model progressively builds upon the existing scene's geometry by conditioning on the rendered keyframe points. This design enhances both the appearance and geometric coherency between the new and previously generated keyframes, thereby maintaining quality and consistency across a long generation horizon.

\begin{figure}[!t]
    \centering
    \includegraphics[width=\linewidth]{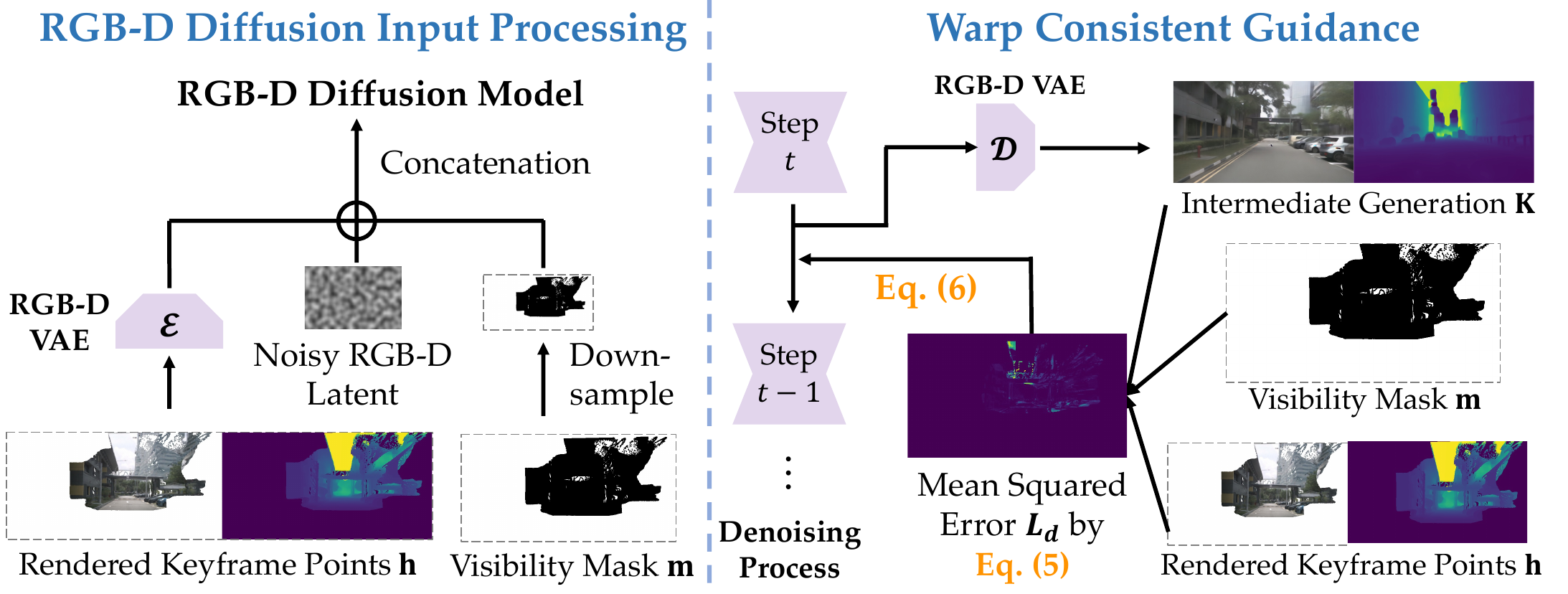}
    \vspace{-2em}
    \caption{(Left) Input processing of our RGB-D diffusion model. (Right) The warp-consistent guidance for steering the sampling process toward better geometric consistency.}
    \vspace{-1.5em}
    \label{fig:method_second}
\end{figure}

\mypara{Backbone} Our RGB-D diffusion model is based on the Latent Diffusion Model (LDM)~\cite{rombach2022ldm}. It comprises a Variational Autoencoder (VAE)~\cite{kingma2013auto} that compresses inputs into a latent space, where the denoising U-Net~\cite{ronneberger2015u} is trained to revert the forward process. To extend the model for RGB-D conditioning and generation, we first extend the VAE to jointly model image and depth, encoding and decoding the RGB-D data to or from the latent space, respectively. Full details of the RGB-D VAE are in \S~\ref{sec:rgbd_vae} of the Appendix. 

\mypara{Conditioning} 
To integrate the Rendered Keyframe Points $\mathbf{h}$ as a condition, as shown in \autoref{fig:method_second} (left), it is first encoded by the RGB-D VAE, serving as an additional conditioning input to the model. The visibility mask $\mathbf{m}$ is down-sampled to match the spatial resolution of the latent space. The noisy latent, the mask, and the RGB-D latent code of $\mathbf{h}$ are concatenated along the channel dimension and fed into the U-Net. To accommodate the additional channels, the U-Net architecture is extended by adding five extra input channels. The initial convolution layers for processing these new channels are zero-initialized. Following Stable Diffusion~\cite{rombach2022ldm}, text prompts are injected through the cross-attention module. We also incorporate HD maps and object boxes through ControlNet~\cite{zhang2023adding}; details are discussed in \S~\ref{sup_sec:setting}.

\mypara{Training Pipeline} The training of the RGB-D diffusion model comprises two stages: RGB-D pre-training and rendering-conditioned fine-tuning. The RGB-D pre-training stage performs the RGB-D inpainting task on large-scale, curated image data. The depth used for training is pseudo-labeled with Metric3D~\cite{hu2024metric3d}, an off-the-shelf monocular metric depth estimator. Note that the additional conditions are simulated by masking the data with synthetic masks, rather than being derived from a keyframe, which is a key for large-scale training on diverse data sources.
In the rendering-conditioned fine-tuning stage, the rendered keyframe points and visibility mask are derived from the last keyframe, while the map and bounding box conditions are also incorporated. The model is fine-tuned on driving-specific datasets.
More training details are provided in \S~\ref{subsec:implementation}.

\subsection{Warp Consistent Guidance}
\label{sec:warpConsistent}

Although diffusion models conditioned on rendered keyframe points (\ie, coarse image) share similarities with traditional image inpainting, we observe that the generated content sometimes exhibits pronounced appearance and geometry inconsistencies in the overlapping regions. 
One potential cause is the noisy training data, where the depths of two consecutive keyframes do not align perfectly with each other. The inconsistency adversely affects 3D consistency, leading to increasingly noticeable shifts in appearance and scene geometry throughout the iterative generation process.

To mitigate this, we propose \textit{Warp Consistent Guidance} (WCG) to steer the sampling process of the diffusion model toward better geometric consistency. The idea is to introduce a projection consistency loss to quantify the discrepancy between rendered keyframe points and RGB-D generation, as illustrated in \autoref{fig:method_second} (right).
The loss then adjusts the sampling process through classifier guidance. Concretely, the loss is defined as the masked Mean Squared Error (MSE) between the predicted RGB-D frame \( \mathbf{x} \) and the rendered keyframe points input \( \mathbf{h} \):
\vspace{-0.5em}
\begin{equation}
    \mathcal{L}_{d}(\mathbf{x},\mathbf{h};\mathbf{m}) = \frac{\sum_{i} \mathbf{m}_i \left( \mathbf{x}_i - \mathbf{h}_i \right)^2}{\sum_{i} \mathbf{m}_i}.
    \label{equ:warpConsistentLoss}
\end{equation}
where $i$ is the pixel index, and we slightly abuse the subscript of \( \mathbf{x} \) here. \( \mathbf{m}_i \in \{0,1\} \) is the \( i \)-th pixel of the overlap mask \( \mathbf{m} \). \( \mathbf{m}_i = 1 \) means the pixel is visible in both the target keyframes and the previously generated keyframes, and \( \mathbf{m}_i = 0 \) otherwise.
We mask out 5\% pixels with the largest loss for better robustness against noise. 

\begin{figure}[!t]
    \centering 
    \includegraphics[width=\columnwidth]{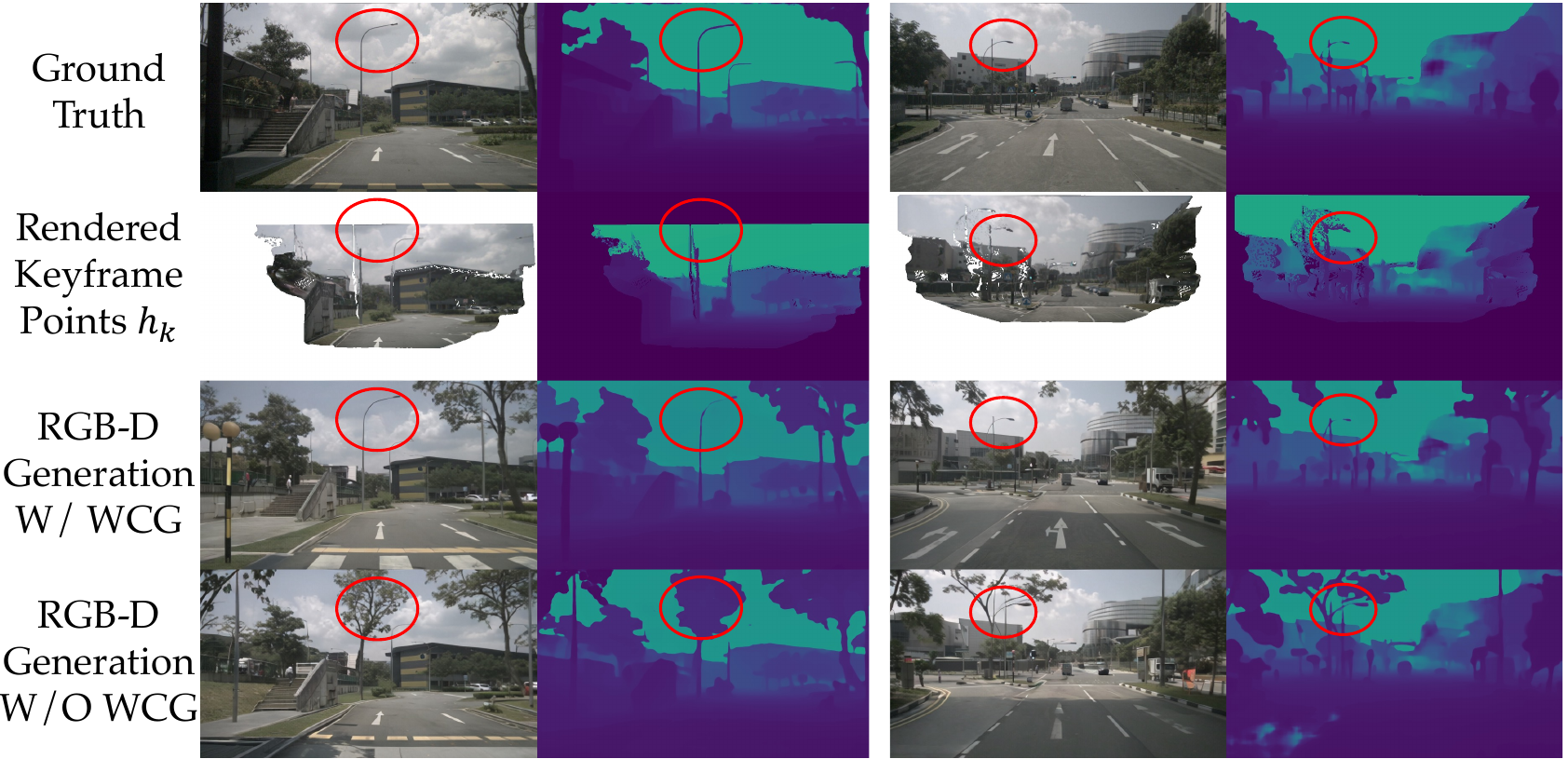}
    \vspace{-1.8em}
    \caption{Qualitative comparison to show the effectiveness of our warp consistent guidance (WCG) strategy for better consistency.}
    \label{fig:ablation_method}
\end{figure}

The gradient of \( \mathcal{L}_{d} \) then guides the generation towards latent regions that are more geometrically consistent with existing keyframes. 
Formally, at timestep \( t \), the sampling process is modified by adjusting the original score estimate \( \mathbf{s}_\theta(\mathbf{x}_t, t) \) with \( \nabla_{\mathbf{x}_t}\mathcal{L}_{d} \). The adjusted score is defined as:
\begin{equation}
    \tilde{\mathbf{s}}_\theta(\mathbf{x}_t, t) = \mathbf{s}_\theta(\mathbf{x}_t, t) + w \nabla_{\mathbf{x}_t} \mathcal{L}_{d}(\mathbf{x}_t, \mathbf{h}; \mathbf{m}),
\end{equation}
\( \tilde{\mathbf{s}}_\theta(\mathbf{x}_t, t) \) is the adjusted score used at each sampling step, and \( w \) is the scale that controls the strength of WCG.
\autoref{fig:ablation_method} demonstrates that WCG significantly improves the consistency between the rendered keyframe points and the newly generated keyframe, which eventually enhances the quality and stability of the long-horizon generation.

\subsection{Video Diffusion Model for Interpolation}
\label{sec:videodiff}

After generating the sparse RGB-D keyframes along the trajectory in the 3D scene, the interpolation stage connects two consecutive keyframes with dense frames, as illustrated in \autoref{fig:pipeline} (right). Compared to an image inpainting model, employing a video diffusion model enables smoother interpolation due to its strong temporal priors.
Furthermore, the video diffusion model can be anchored on our high-quality RGB-D keyframes. The rendering results provide effective geometric cues, making it easier to produce coherent interpolation.
The interpolation process is defined by:
\begin{equation}
    \{\mathbf{x}^I_t\}_{t=1}^T = G(\{\mathbf{z}_t\}_{t=1}^T; \{\mathbf{h}^I_t\}_{t=1}^T, \mathbf{K}_1, \mathbf{K}_2).
\end{equation}
\( G \) is the video diffusion model, and we use the off-the-shelf point-cloud-conditioned model from ViewCrafter~\cite{yu2024viewcrafter}. $T$ is the number of interpolation frames between two keyframes.
Each \( \mathbf{z}_t \) is sampled from \( \mathcal{N}(\mathbf{0}, \mathbf{I}) \).
\( \{\mathbf{x}^I_t\}_{t=1}^T \) are the output frames and
\(\{\mathbf{h}^I_t\}_{t=1}^T \) are the corresponding rendered keyframe points. Unlike the rendered results $\mathbf{h}$ used in keyframe generation. \(\mathbf{h}^I \) only contains the RGB information without depth to fit the pre-trained model's input specification. \( \mathbf{K}_1 \) and \( \mathbf{K}_2 \) are the two consecutive RGB-D keyframes.

\begin{table*}[!th]
\centering
\caption{Comparison of Methods in Terms of FID and FVD at Different Time Splits. WonderJourney$^\dag$ indicated WonderJourney adapted for driving scene. Overall indicates the time split of 0-20s. For both FID and FVD, lower is better, denoted by $\downarrow$. The top-performing methods are highlighted in \textbf{bold}. The proposed method advances the previous State-of-the-Art method Vista~\cite{gao2024vista} by reducing FID from 68.3 to 35.1 and FVD from 629.8 to 359.0, corresponding to a significant improvement margin of 48.6\% and 43.0\%, respectively.  
\vspace{-0.7em}
}
\label{tab:comparison}
\begin{tabularx}{\textwidth}{l *{10}{>{\centering\arraybackslash}X}}
\toprule
\multirow{2}{*}{Method} & \multicolumn{2}{c}{0--5s} & \multicolumn{2}{c}{5--10s} & \multicolumn{2}{c}{10--15s} & \multicolumn{2}{c}{15--20s} & \multicolumn{2}{c}{Overall} \\
\cmidrule(lr){2-3} \cmidrule(lr){4-5} \cmidrule(lr){6-7} \cmidrule(lr){8-9} \cmidrule(lr){10-11}
& FID$\downarrow$ & FVD$\downarrow$ & FID$\downarrow$ & FVD$\downarrow$ & FID$\downarrow$ & FVD$\downarrow$ & FID$\downarrow$ & FVD$\downarrow$ & FID$\downarrow$ & FVD$\downarrow$ \\
\midrule
WonderJourney~\cite{yu2024wonderjourney}            &  93.0             &  977.2        & 157.4         & 1651.7        &  172.7        & 1716.7        &   172.5       & 1737.8        &  127.5        &  1017.4       \\
WonderJourney$^\dag$~\cite{yu2024wonderjourney}     &  49.8             &   661.8       & 111.7         &  1551.5       &  114.1        & 1730.1        &   99.0        &  1756.7       &  73.7         &   939.6       \\
Vista~\cite{gao2024vista}                           &  37.2             &   436.4       & 72.4          &  967.0        &  124.6        & 1329.1        &   157.9       &  1614.5       &  68.3         &   629.8       \\
\ourmethod (Ours)                                 &  \textbf{34.3}    & \textbf{385.9}& \textbf{48.8} & \textbf{526.3}&\textbf{54.0}   &\textbf{579.5}&\textbf{56.8}  &\textbf{657.4} & \textbf{35.1} &\textbf{359.0} \\
\bottomrule
\end{tabularx}
\end{table*}

\section{Experiment}
\label{sec:exp}

\mypara{Baselines} We compare \ourmethod with Vista~\cite{gao2024vista} and WonderJourney~\cite{yu2024wonderjourney}, two competitive methods for generating long-horizon videos. Vista is the state-of-the-art in driving video generation. WonderJourney achieves long-horizon generation for general scenes. We adapt WonderJourney by finetuning its diffusion model on driving data and performing the generation with vehicle trajectories from nuScenes (denoted as WonderJourney$^\dagger$) for fair comparisons.

\mypara{Datasets and Evaluation Metrics} We evaluate all methods on the nuScenes validation set (with 150 videos) using Fréchet Inception Distance (FID)~\cite{heusel2017gans} and Fréchet Video Distance (FVD)~\cite{unterthiner2018towards} while extending the evaluation to focus more on \textbf{long horizon}. Each model generates a long sequence of frames from a single input image, and the generation can extend up to 20 seconds. To better understand the performance change over time, we compute FID and FVD scores over consecutive 5‑second segments.

In the rest of this section, \S\ref{subsec:implementation} introduces the primary implementation details of \ourmethod, then \S\ref{subsec:exp_main}, \S\ref{subsec:ablation} and \S\ref{subsec:more_results} present the evaluation results and analyses. We refer to the supplementary for more details and results.

\subsection{Implementation Details}
\label{subsec:implementation}

\mypara{RGB-D Pre-training} RGB-D pre-training scales the diffusion model on large-scale RGB-D datasets to learn robust geometry priors. While there are many existing high-quality image datasets~\cite{schuhmann2022laion,changpinyo2021conceptual-12m}, depth data is scarce. To scale up the training, we generate depth with a monocular metric depth predictor~\cite{yin2023metric3d,hu2024metric3d}. In practice, we collect RGB images from nuScene (training split)~\cite{caesar2020nuscenes}, Argoverse2 (training split)~\cite{wilson2023argoverse2}, and SA1B~\cite{kirillov2023segment-anything} and prepare the depth data, forming a dataset of 13 million diverse images. We use the ground-truth intrinsics for the depth predictor~\cite{hu2024metric3d} on nuScene and Argoverse2, while predicting the intrinsics with WildCamera~\cite{zhu2024tame} on SA1B. We also generate text pseudo-labels with a Vision-Language Model~\cite{lin2024vila} for pretraining.
We initialize the Diffusion Unet with the pre-trained Unet of SD-Inpainting-V2.0~\cite{rombach2022ldm}. We train the model with text-conditioned RGB-D in-painting to preserve the text controllability and inpainting ability of the base model. The inpainting masks are randomly generated to resemble the visibility mask from projection.
We use 32 A100 GPUs for RGB-D pre-training, training for 50k iterations over 2 days. The batch size is 1024 and the learning rate is 1e-4.

\mypara{Rendering Conditioned Training} For rendering conditioned training, we employ a training strategy mimicking the iterative keyframes generation process. Specifically, each training sample is generated via sampling a pair of frames from the same video sequence with a gap range from 5 to 60 frames. Assigning one of them to be the condition frame and the other as the target, we then project the condition frame to the target frame utilizing the depth and cameras. The projection then serves as the rendered keyframe points conditioning for the target frame. The model is also conditioned with HD maps and object boxes. The training only uses nuScenes, and we prepare 500 samples per scene, resulting in a dataset with 350k samples. The training is conducted on 8 A6000 GPUs over 2 days, with a batch size of 512 for 20k iterations, using a learning rate of 1e-4.

\mypara{Data Filtering} The rendered keyframe points are sometimes inconsistent with the target image due to noisy depth, dynamic objects, and occlusions. This can impair the 3D consistency of the iterative generation. To alleviate this, we use the warp consistent loss \( \mathcal{L}_{d}(x,h;m) \) to measure the consistency between two frames and filter out the most inconsistent samples. In practice, we filter out 20\% of the 350k samples and train only with the remaining 280k.

\begin{figure*}[!th]
\centering
\includegraphics[width=\textwidth]{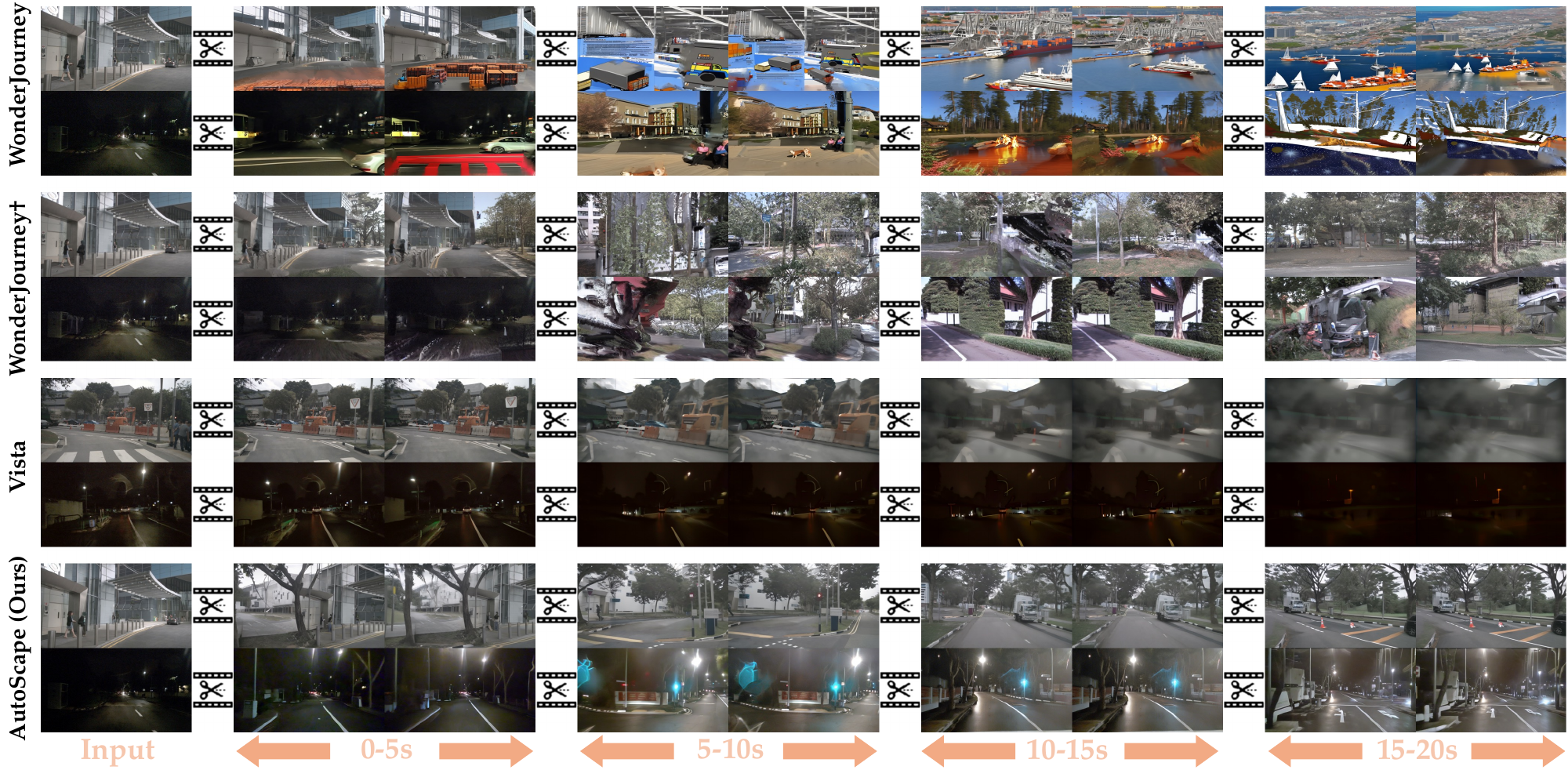}
\vspace{-2em}
\caption{Qualitative comparisons. WonderJourney$^\dag$ represents WonderJourney adapted for driving scenes. We sample frames from different time splits to demonstrate both short and long-term performance. \ourmethod maintains 3D consistency over significant view changes and extends toward a long horizon. Vista generates the video from the start of the trajectory to the end, while others generate in the reverse order. 
}
\vspace{-1em}
\label{fig:mainExp}
\end{figure*}

\mypara{Video Diffusion Model Training Setting} 
Given the high-quality RGB-D keyframes, the pre‑trained model from ViewCrafter~\cite{yu2024viewcrafter} can directly produce promising interpolation results. We therefore use the point-cloud-conditioned model without fine‑tuning. Several concurrent methods~\cite{ren2025gen3c,bai2025recammaster,mark2025trajectorycrafter} are also potentially applicable to this stage, and we leave systematic comparisons for future work.

\mypara{Viewpoints Selection for Keyframes} Selecting an optimal spacing for keyframes is essential. On the one hand, overly dense keyframes result in inefficient modeling of long-range geometric dependencies. On the other hand, if the keyframes are too sparse, the interpolation stage could fail. We designate the first keyframe as one endpoint of the trajectory, then traverse the trajectory to identify the subsequent keyframes. The first viewpoint with either the distance or the view angle difference from the previous keyframe exceeding \( \beta \) or \( \gamma \), respectively, is selected as the next keyframe. In practice. We set \( \beta = 10 \, m \) and \( \gamma = 20^\circ \).

\subsection{Main Results} 
\label{subsec:exp_main}

\mypara{Quantitative Comparison} \autoref{tab:comparison} shows that \ourmethod achieves the best quantitative performance across all time splits. For overall video quality, it improves Vista from [68.3, 629.8] to [35.1, 359.0] in terms of [FID, FVD], respectively. This margin is even larger in terms of the long-horizon generation in time split of 15-20s, from [157.9, 1614.5] to [56.8, 657.4]  in terms of [FID, FVD], corresponding to a significant improvement of  64\% and 59.3\%.
To evaluate generalizability, we further assess our method on the Argoverse2 dataset~\cite{wilson2023argoverse2} without fine-tuning. While Vista achieved [FID, FVD] scores of [80.4, 614.2], \ourmethod significantly improves the scores to [49.2, 317.9].
demonstrating the robust generalizability achieved through the RGB-D pre-training.

\mypara{Qualitative Comparison} As visualized in \autoref{fig:mainExp}, while WonderJourney produces reasonable images across all time splits, it struggles to generate photorealistic driving scenes. After adapting it to driving data, WonderJourney$^\dagger$ shows better photorealism. However, both versions exhibit a gradual loss of context and deviate from the input image. For instance, in the video clip spanning 5–10s, both versions alter the scene from night to day. Additionally, WonderJourney depends heavily on carefully designed camera trajectories, and the results deteriorate with real-world vehicle trajectories. Vista, on the other hand, can generate high-quality videos for short clips (\eg, 0–5 seconds), but the performance rapidly drops for more extended sequences. In contrast, \ourmethod consistently produces 3D-consistent, long-horizon, high-quality videos across all temporal splits, effectively handling significant dynamics of real-world trajectories and generalizing well to challenging conditions

\mypara{User Study} To thoroughly evaluate long-sequence 3D consistency, we conducted a user study in which participants selected the video exhibiting the best 3D consistency performance over a long sequence. From 22 valid responses, our method was preferred in 88.39\% of the cases against Vista and both variants of WonderJourney.

\begin{table}[!t]
\centering
\caption{Ablation study on the design components of \ourmethod.  
}
\vspace{-0.5em}
\label{tab:ablationStudy}
\begin{tabularx}{\columnwidth}{l *{2}{>{\centering\arraybackslash}X}}
\toprule
\textbf{Methods} & FID$\downarrow$ & FVD$\downarrow$ \\
\midrule
\ourmethod (Ours)                        &\textbf{35.1}   &\textbf{359.0} \\
-- RGB-D Pre-training          &    47.6        &  650.0        \\
-- Data Filtering             &    43.5        &  463.0        \\
-- Warp Consistent Guidance   &    38.5        &  380.2        \\
-- Depth Generation           &    39.2        &  511.4        \\
\bottomrule
\vspace{-2.4em}
\end{tabularx}
\end{table}

\begin{figure*}[t]
  \centering
\includegraphics[width=\textwidth]{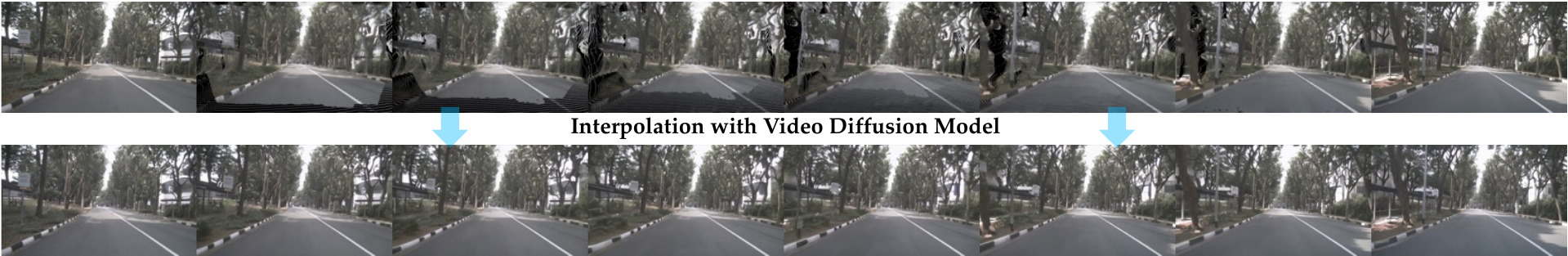}
\vspace{-1.7em}
\caption{Illustration of the interpolation process. (Top) The intermediate frames between two RGB-D keyframes with the rendered interpolation points. 
(Bottom) The corresponding interpolation results from the video diffusion model. 
}
\vspace{-0.5em}
\label{fig:render}
\end{figure*}

\begin{figure*}[t]
  \centering
\includegraphics[width=\textwidth]{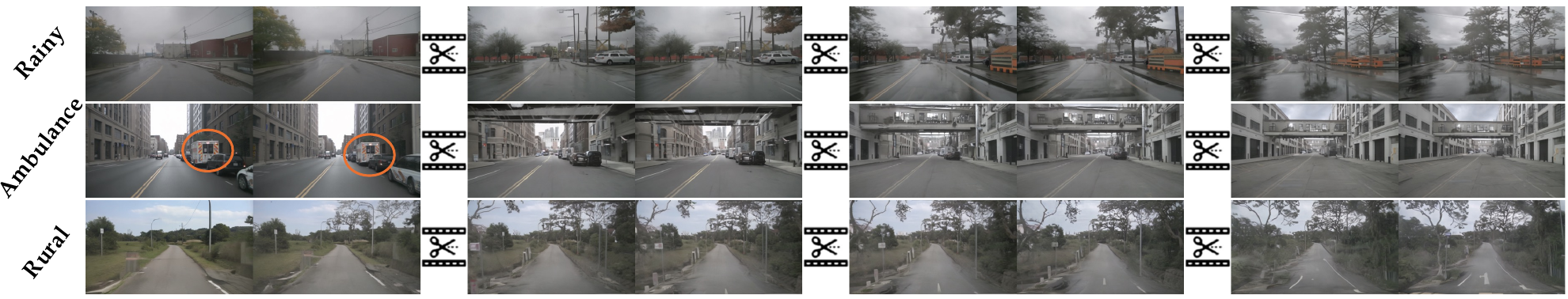}
\vspace{-2em}
\caption{Generalization to various weather conditions, rare objects, and uncommon scene types.
}
\vspace{-1em}
\label{fig:corner_case}
\end{figure*}

\subsection{Ablation Study} 
\label{subsec:ablation}

We investigate the effectiveness of different components of \ourmethod in \autoref{tab:ablationStudy} and analyze the results below.

\mypara{RGB-D Pretraining} For this ablation, instead of employing the two-stage training pipeline, we initialize the model with the Stable Diffusion model and finetune it with only the Rendering Conditioned Training stage. Removing the pre-training stage significantly impacts performance, increasing the FID and FVD from 35.1 and 359.0 to 47.6 and 650.0, respectively, highlighting the effectiveness of large-scale RGB-D pre-training.

\mypara{Data Filtering} As introduced earlier, we filter out 20\% most inconsistent samples for training. Training with noisy data results in significant performance degradation of 8.4 and 104 in FID and FVD, respectively. This indicates that the inconsistency between the rendered keyframe points and the generated content has a significant impact on the consistency. 

\mypara{Warp Consistent Guidance} Warp consistent guidance can significantly boost the FID and FVD of the generated video by further enhancing the 3D consistency. Removing it would result in a performance drop from 35.1 to 38.5 for FID and from 359.0 to 380.2 for FVD, indicating the effectiveness of the proposed Warp Consistent Guidance.

\mypara{Joint Depth Generation} We then study the necessity of jointly generating depth with the diffusion model. For this ablation study, we replace the RGBD Diffusion model with the standard RGB Diffusion model. The depth is predicted using a monocular metric depth prediction model~\cite{hu2024metric3d} based on the generated RGB image. As shown in \autoref{tab:ablationStudy}. Removing depth generation decreases the FID and FVD from 35.1 and 359.0 to 39.2 and 511.4, respectively. The clear degradation of FVD indicates that the model without depth generation lacks an understanding of geometry and leads to worse 3D consistency.

\begin{figure}[!t]
    \centering
    \includegraphics[width=\columnwidth]{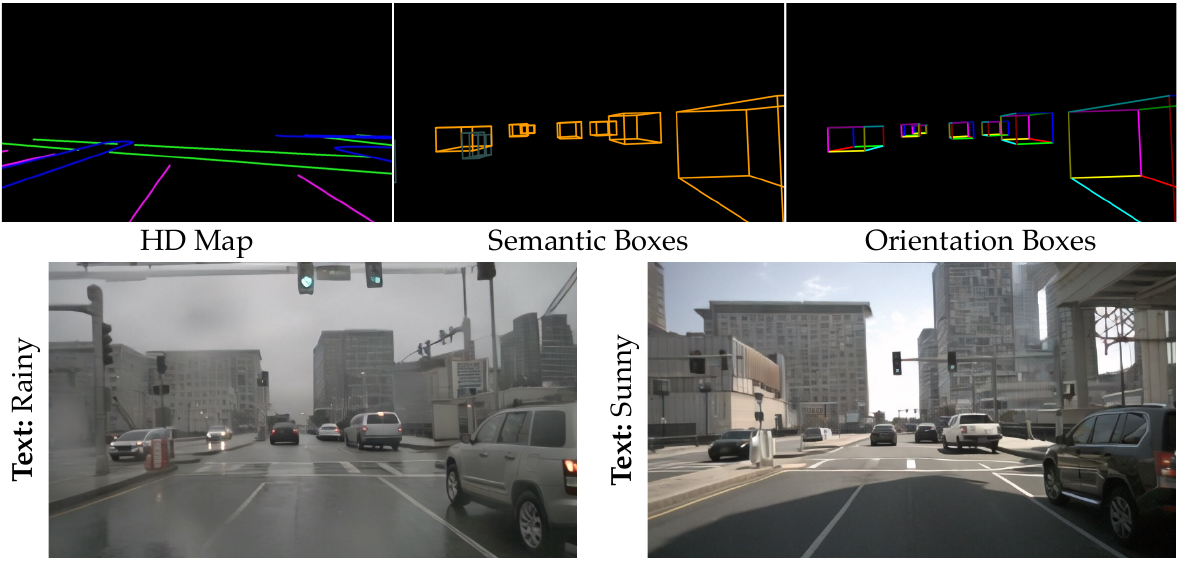}
    \vspace{-1.7em}
    \caption{\ourmethod's generation controllability with texts (for weather), HD maps, and object bounding boxes.  
    }
    \vspace{-1.5em}
    \label{fig:control}
\end{figure}

\subsection{More Results and Analyses}
\label{subsec:more_results}

\mypara{Interpolation Process}
\autoref{fig:render} illustrates the process of interpolation generation with the intermediate rendering condition. The projection of point clouds depicts the appearance of the global frame, ensuring the 3D consistency of generated content. And the video diffusion model complements the missing holes and produces a high-quality video clip. More examples can be found in the supplementary material.

\mypara{Generalization to Corner Case} 
\autoref{fig:render} illustrates how \ourmethod handles out-of-distribution (OOD) corner cases, including rare weather conditions (\eg, rainy), uncommon vehicle types (\eg, ambulances), and uncommon environments (\eg, rural landscapes). \autoref{fig:mainExp} also covers the generation of night-time scenes.

\mypara{Controllability} \autoref{fig:control} demonstrates the fine-grained controllability of our method through texts, object bounding boxes, and HD maps. The flexible control enables the generation of highly customized, long-horizon scenes.

\section{Conclusion}
\label{sec:conclusion}

This paper introduced \ourmethod, a hierarchical framework for long-horizon, 3D-consistent driving scene generation. The core is a novel RGB-D diffusion model that generates geometrically-consistent keyframes using joint RGB-D modeling, explicit geometry conditioning, and warp-consistent guidance. By anchoring a video interpolation stage with high-quality keyframes, \ourmethod successfully mitigates geometric drift, generating realistic and coherent driving scene videos that maintain consistency for over 20 seconds.

\clearpage
{
    \small
    \bibliographystyle{ieeenat_fullname}
    \bibliography{main}
}

\clearpage
\setcounter{page}{1}
\maketitlesupplementary
\appendix

In the supplementary material, we provide additional content that could not be included in the main paper due to page and format constraints. The supplementary material is organized as follows:
\begin{itemize}

\item In \S~\ref{sup_sec:setting} presents the remaining implementation details.

\item In \S~\ref{sec:rgbd_vae} presents the architectural and training details of the RGB-D VAE of \ourmethod.

\item In \S~\ref{sec:supp:results} provides additional experimental results.

\end{itemize}

\section{Remaining Implementation Details}
\label{sup_sec:setting}

This section presents the remaining implementation details that are not covered in the main paper due to space limitations. The proposed method is implemented with Pytorch~\cite{paszke2019pytorch} and the Diffuser library.

\mypara{Optimization Settings} For both RGB-D pretraining and rendering-conditioned training, we utilize the AdamW optimizer to facilitate optimization. The learning rate (\texttt{lr}) and weight decay (\texttt{wd}) are set to \(1 \times 10^{-4}\) and \(1 \times 10^{-2}\), respectively, with a learning rate warmup applied over the first 3000 iterations. Gradient clipping with a maximum norm of 1 is implemented to enhance training stability. Additionally, both training and inference are conducted using bfloat16 (brain floating-point 16-bit) precision to ensure computational efficiency and optimization effectiveness.

\begin{figure}[!t]
    \centering
    \includegraphics[width=\columnwidth,trim={0cm 9cm 0cm 0cm}, clip]{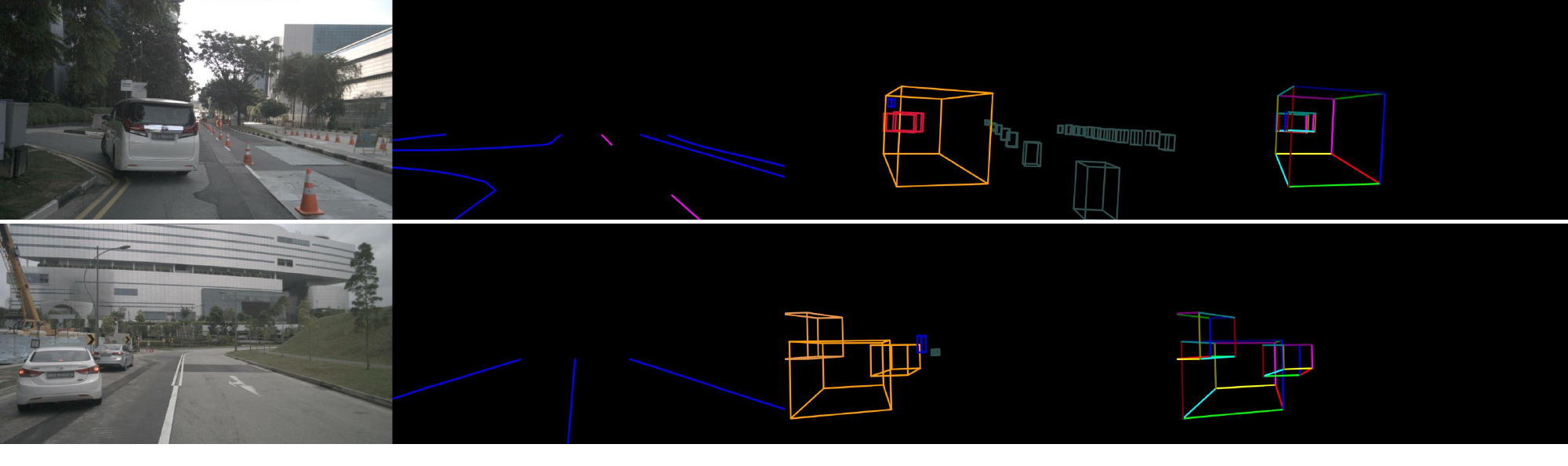}
    \vspace{-4mm}
    \caption{The control signals with the corresponding images. From left to right are ground-truth RGB images, projected maps, semantic box control, and orientation box control.
    \vspace{-3mm}
    }
    \label{fig:cond_vis}
\end{figure}

\mypara{HD Map and Bbox Condition}  To enable more flexible controllability, we augment our RGB-D diffusion model with a ControlNet~\cite{zhang2023adding} branch to encode HD maps and object bounding boxes. \autoref{fig:cond_vis} provides a visualization of these conditioning inputs. Specifically, for the map condition, we extract the layers (\ie, lane boundary, lane divider, and pedestrian crossings from the vector HD maps~\cite{li2022hdmapnet,yuan2024streammapnet,chen2024maptracker} and then project them onto the image plane. To specify the location and orientation of objects precisely, we utilize two types of box control images: semantic box control and orientation box control. Both box controls are derived by projecting 3D bounding boxes onto the image plane with the camera parameters. For the semantic box control, different colors are used to distinguish vehicles, pedestrians, roadblocks, etc. For the orientation box control, the orientation of each vehicle is indicated by assigning unique colors to each edge of the box. \autoref{fig:control} in the main paper demonstrates the controlled generation through these protocols. Note that our conditioning strategies for HD Maps and objects are different from those in MagicDrive~\cite{gao2023magicdrive} or DriverDreamer~\cite{wang2023drivedreamer}.

\mypara{Training with ControlNet} ControlNet is only incorporated during the rendering-conditioned training stage, as the HD maps and object boxes conditions are not available for the RGB-D pre-training stage, where we use datasets beyond driving. The ControlNet is initialized using the U-Net model from the RGB-D pretraining stage, following those outlined in the original ControlNet~\cite{zhang2023adding}. During the rendering-conditioned training stage, we fine-tune both ControlNet and U-Net to facilitate convergence.

\mypara{Inference Settings} For diffusion model inference, we utilize DPM-Solver~\cite{lu2022dpm} with 50 steps. Additionally, classifier-free guidance~\cite{ho2022classifier} is employed to enhance the quality of conditioned generation, using a guidance strength of 7.5 in accordance with the default settings of the diffuser library.

\section{Details of the RGB-D VAE}
\label{sec:rgbd_vae}
Similar to LDM3D~\cite{stan2023ldm3d}, we modify the VAE to support depth encoding and decoding to accommodate depth generation, while preserving the latent code shape. Specifically, we first normalize the depth to 0-1, with a maximum depth of 300 meters, to align with the scale of the RGB channels. Then, the normalized depth (1 channel) is concatenated with RGB (3 channels) to create a 4-channel RGB-D input for the VAE. Architecturally, we extend the first and last convolutions in both the encoder and decoder to accommodate this 4-channel input and output, ensuring compatibility with RGB-D data. As the default 8-bit choice for RGB channel leads to significant precision loss for depth channel~\cite{stan2023ldm3d}, we employ 16-bit precision for RGB-D inputs and outputs to retain depth details accurately. Since the latent feature shape remains unchanged, we apply the existing U-Net architecture directly for latent diffusion.

The RGB-D VAE is initialized with the pretrained RGB VAE from Stable Diffusion models~\cite{rombach2022ldm}. The added parameters are initialized to zero to preserve pretrained knowledge. The optimization target is defined as 
\begin{equation}
\begin{split}
    \mathcal{L}_{\text{VAE}} = &\;  \mathbb{E}_{q_\phi(\mathbf{z} \mid \mathbf{x})} \left[ -\log p_\theta(\mathbf{x}_{\text{rgb}} \mid \mathbf{z}) \right] \\
    &+ \lambda_{\text{depth}} \cdot \mathbb{E}_{q_\phi(\mathbf{z} \mid \mathbf{x})} \left[ -\log p_\theta(\mathbf{x}_{\text{depth}} \mid \mathbf{z}) \right] \\
    &+ D_{\text{KL}} \left( q_\phi(\mathbf{z} \mid \mathbf{x}) \,\|\, p(\mathbf{z}) \right)
\end{split}
\end{equation}
where $ \mathbf{x}_{\text{rgb}} $ represents the RGB image data. $ \mathbf{x}_{\text{depth}} $ represents the depth map data. $\mathbf{x}$ is the combination of $ \mathbf{x}_{\text{rgb}} $ and $ \mathbf{x}_{\text{depth}} $.
$q_\phi(\mathbf{z}\mid \mathbf{x}_{\text{rgb}}, \mathbf{x}_{\text{depth}}) $ is the encoder network with parameters $\phi$, encoding both RGB and depth inputs.
$p_\theta(\mathbf{x}_{\text{rgb}}\mid \mathbf{z})$ and $p_\theta(\mathbf{x}_{\text{depth}} \mid \mathbf{z})$ are the decoder networks reconstructing RGB images and depth maps from the latent variable $\mathbf{z}$.  $D_{\text{KL}}$ is the Kullback-Leibler divergence between the approximate posterior $q_\phi(\mathbf{z} \mid \mathbf{x})$ and the prior $p(\mathbf{z}) $.

The first and second term, $\mathbb{E}_{q_\phi(\mathbf{z} \mid \mathbf{x})} \left[ -\log p_\theta(\mathbf{x}_{\text{rgb}} \mid \mathbf{z}) \right]$ and $\mathbb{E}_{q_\phi(\mathbf{z} \mid \mathbf{x})} \left[ -\log p_\theta(\mathbf{x}_{\text{depth}} \mid \mathbf{z}) \right]$, minimize the reconstruction errors for the RGB images and depth maps, respectively. The third term $D_{\text{KL}} \left( q_\phi(\mathbf{z} \mid \mathbf{x}) \,\|\, p(\mathbf{z}) \right)$, regularizes the latent space by enforcing alignment with a predefined prior distribution, thereby promoting smoothness and continuity in the latent space $\mathbf{z}$.

Given that depth maps tend to contain less high-frequency information than RGB images due to the inherently smooth nature of geometric data, the reconstruction loss for depth is generally smaller than for RGB. To address this imbalance, we introduce a weighting factor, $\lambda_{\text{depth}}$, to amplify the depth reconstruction loss. In practice, we set $\lambda_{\text{depth}} = 10$. 

To train the RGB-D diffusion model, we implement a two-stage training strategy, as outlined in \S~\ref{subsec:implementation}.

\section{Additional Experimental Results}
\label{sec:supp:results}

\mypara{More baseline results} To further evaluate the quality of keyframes generated by the proposed \ourmethod in comparison to the baselines, we apply ViewCrafter to interpolate the keyframes
produced by WonderJourney$^\dagger$. This results in FID and FVD scores of 59.1 and 858.9, respectively, which are significantly higher than those achieved by \ourmethod (35.1 and 359.0). These findings highlight the superior visual quality of the keyframes generated by our method.

\mypara{Compare with single-stage video diffusion models} 
To further assess the performance of our proposed two-stage method against the state-of-the-art one-stage approach, we fine-tune COSMOS-Transfer~\cite{alhaija2025cosmos} with the HD map from nuScenes and perform autoregressive generation to produce long videos. COSMOS‑Transfer achieved an FID of 44.2 and an FVD of 436.1, whereas our method attained 35.1 and 359.0, respectively, demonstrating its clear superiority.

\end{document}